\documentclass[11pt]{article}
\usepackage{acl2016}
\usepackage{times}
\usepackage{latexsym}

\aclfinalcopy 

\usepackage{tikz}
\usetikzlibrary{bayesnet}

\usepackage{multirow}
\usepackage{caption}
\usepackage{subcaption}
\usepackage[singlelinecheck=false]{caption}

 \usepackage{amsmath, amsthm, amssymb}

\usepackage{enumitem}

\title{A Trolling Hierarchy in Social Media and \\A Conditional Random Field For Trolling Detection}

\author{Luis Gerardo Mojica\\
  The University of Texas at Dallas\\
  Richardson, TX 75083-0688\\
  luis.mojica@utdallas.edu}

\begin{document}

\maketitle

\begin{abstract}
An-ever increasing number of social media websites, electronic newspapers and Internet forums allow visitors to leave comments for others to read and interact. This exchange is not free from participants with malicious intentions, which do not contribute with the written conversation. Among different communities users adopt strategies to handle such users. In this paper we present a comprehensive categorization of the trolling phenomena resource, inspired by politeness research and propose a model that jointly predicts four crucial aspects of trolling: intention, interpretation, intention disclosure and response strategy. Finally, we present a new annotated dataset containing excerpts of conversations involving trolls and the interactions with other users that we hope will be a useful resource for the research community.

\end{abstract}

\section{Introduction}

In contrast to traditional content distribution channels like television, radio and newspapers, Internet opened the door for direct interaction between the content creator and its audience. One of these forms of interaction is the presence of comments sections that are found in many websites. The comments section allows visitors, authenticated in some cases and unauthenticated in others, to leave a message for others to read. This is a type of multi-party asynchronous conversation that offers interesting insights: one can learn what is the commenting community thinking about the topic being discussed, their sentiment, recommendations among many other. There are some comment sections in which commentators are allowed to directly respond to others, creating a comment hierarchy. These kind of written conversations are interesting because they bring light to the types interaction between participants with minimal supervision. This lack of supervision and in some forums, anonymity, give place to interactions that may not be necessarily related with the original topic being discussed, and as in regular conversations, there are participants with not the best intentions. Such participants are called \emph{trolls} in some communities.

Even though there are some studies related to trolls in different research communities, 
there is a lack of attention from the NLP community. We aim to reduce this gap by presenting a comprehensive categorization of trolling and propose two models to predict trolling aspects. First, we revise the some trolling definitions: ``Trolling is the activity of posting messages via communication networks that are in tended to be provocative, offensive or menacing'' by \cite{bishop2013effect}, this definition considers trolling from the most negative perspective where a crime might be committed. In a different tone, \cite{hardaker2010trolling} provides a working definition for \emph{troll}: ``A troller in a user in a computer mediated communication who constructs the identity of sincerely wishing to be part of the group in question, including professing, or conveying pseudo-sincere intentions, but whose real intention(s) is/are to cause disruption and/or trigger or exacerbate conflict for the purpose of their own amusement''. 
These definitions inspire our trolling categorization, but first, we define a \emph{trolling event}: a comment in a conversation whose intention is to cause conflict, trouble; be malicious, purposely seek or disseminate false information or advice; give a dishonest impression to deceive; offend, insult, cause harm, humiliation or aggravation. Also, a troll or troller is the individual that generates a trolling event, trolling is the overall phenomena that involves a troll, trolling event and generates responses from others. Any participant in a forum conversation may become a troll at any given point, as we will see, the addressee of a trolling event may choose to reply with a trolling comment or counter-trolling, effectively becoming a troll as well.


We believe our work makes four contributions.
First, unlike previous computational work on trolling, which focused primarily on analyzing the narrative retrospectively by the victim (e.g., determining the trolling type and the role played by each participant), we study trolling by analyzing comments in a conversation, aiming instead to identify trollers, who, once identified, could be banned from posting.
Second, while previous work has focused on analyzing trolling from the troll's perspective, we additionally model trolling from the target's perspective, with the goal understanding the psychological impact of a trolling event on the target, which we believe is equally important from a practical standpoint.
Third, we propose a comprehensive categorization of trolling that covers not only the troll's intention but also the victim and other commenters' reaction to the troll's comment. We believe such categorization will provide a solid basis on which future computational approaches to trolling can be built.
Finally, we make our annotated data set consisting of 1000 annotated trolling events publicly available. We believe that our data set will be a valuable resource to any researcher interested in the computational modeling of trolling.

\subsection{Trolling Categorization}

Based on the previous definitions we identify four aspects that uniquely define a trolling event-response pair: \textbf{1) Intention}: what is the author of the comment in consideration purpose, a) \emph{trolling}, the comment is malicious in nature, aims to disrupt, annoy, offend, harm or spread purposely false information, b) \emph{playing} the comment is playful, joking, teasing others without the malicious intentions as in a), or c) \emph{none}, the comment has no malicious intentions nor is playful, it is a simple comment. \textbf{2) Intention Disclosure:} this aspect is meant to indicate weather a trolling comment is trying to deceive its readers, the possible values for this aspect are a) the comment's author is a troll and is trying to hide its real intentions, and pretends to convey a different meaning, at least temporarily, b) the comment's author is a troll but is clearly exposing its malicious intentions and c) the comment's author is not a troll, therefore there are not hidden or exposed malicious or playful intentions. There are two aspects defined on the comments that direct address the comment in consideration, \textbf{3) Intentions Interpretation}: this aspect refers to the responder's understanding of the parent's comment intentions. There possible interpretations are the same as the intentions aspect: \emph{trolling}, \emph{playing} or \emph{none}. The last element, is the \textbf{4) Response strategy} employed by the commentators directly replaying to a comment, which can be a trolling event. The response strategy is influenced directly by the responder's interpretation of the parent's comment intention. We identify 14 possible response strategies. Some of these strategies are tied with combinations of the three other aspects. We briefly define each of them in the appendix.

Figure \ref{fig:hierarchy} shows this categories as a hierarchy. Using this trolling formulation, the suspected troll event and the responses are correlated and one cannot independently name the strategy response without learning about the other three aspects. This is challenging prediction problem that we address in this work. 

\begin{figure*}
\centering
  \includegraphics[width=\textwidth]{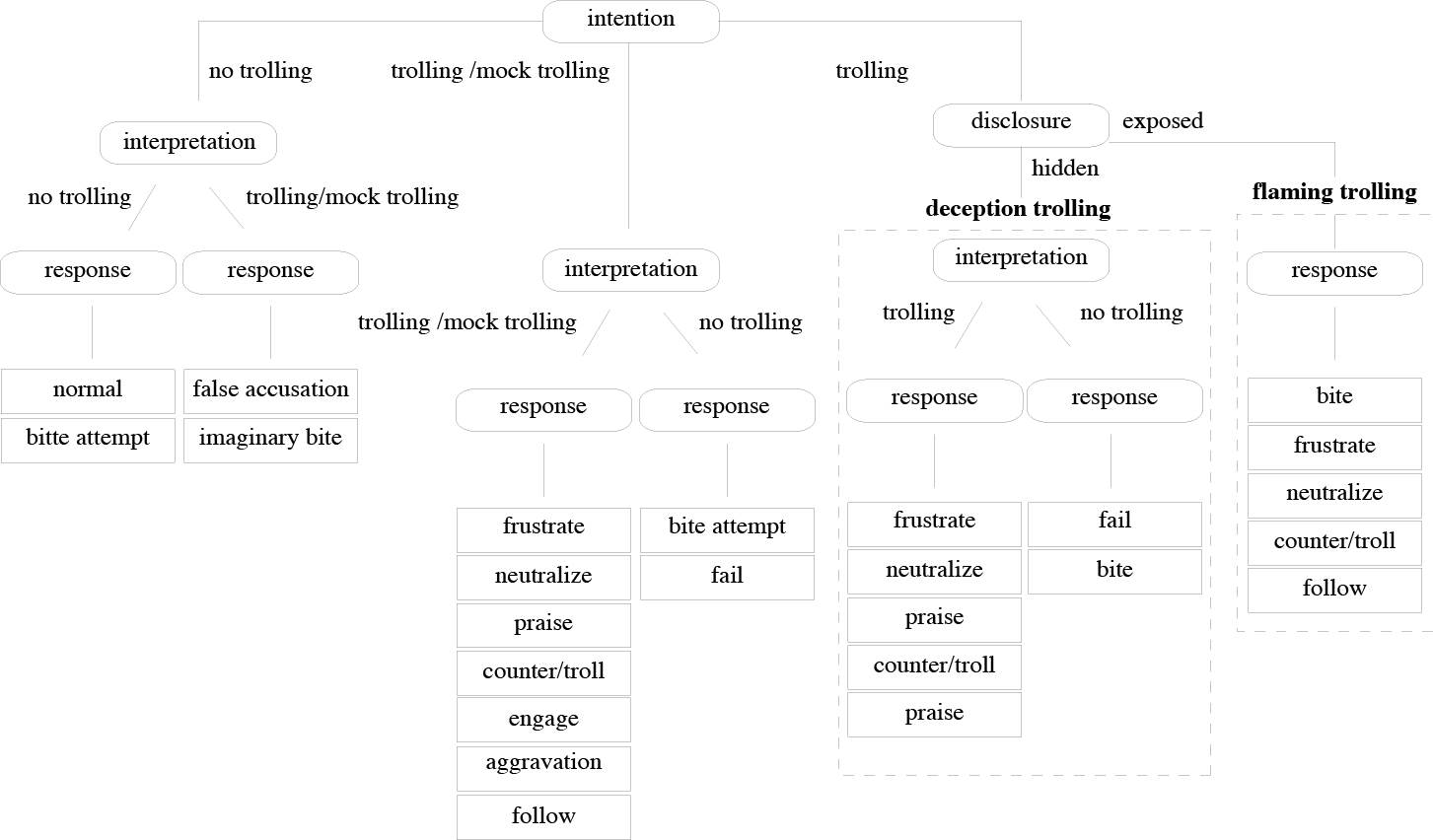}
  \caption{Trolling categorization based on four aspects: Comment's Intention and Intentions Disclosure, and Response's Interpretation and Strategy}
  \label{fig:hierarchy}
\end{figure*}

\subsection{Conversations Excerpts Examples}

To illustrate this hierarchy, we present some examples. These are excerpts from original conversations; the first comment, generated by author \texttt{C0}, on each excerpt is given as a minimal piece of context, the second comment, by the author \texttt{C1} in italics, is the comment suspected to be a trolling event. The rest of the comments, are all direct responses to the suspected trolling comment. When the response author ``name'' is the same as the first comment, it indicates that the that same individual also replied to the suspected troll.

\noindent\textbf{Example 1.}
\vspace{-0.1mm}
\begin{itemize}[noitemsep,nolistsep]
  \small{\item[\texttt{C0:}] My friend who makes \$20,000 a year leased a brand new Chevy Spark EV, for only \$75 per month and he got a California rebate for driving an electric car. Much cheaper than buying older car which usually require heavy upkeep due to its mileage. At this point I think you're just trolling.}
  \begin{itemize}[noitemsep,nolistsep]
    \small{\item [\texttt{\textbf{C1}}:] \emph{IYour friend has a good credit score, which can't be said about actual poor people. Did you grow up sheltered by any chance?}}
    \begin{itemize}[noitemsep,nolistsep]
      \small{\item[\texttt{C0:}] Judging by your post history, you're indeed a troll. Have a good one.}
    \end{itemize}
  \end{itemize}
\end{itemize}

In this example, when \texttt{C1} asks ``Did you grow up sheltered by any chance?", her \emph{intention} is to denigrate or be offensive, and it is not hiding it, instead he is clearly \emph{disclosing} her trolling intentions. In \texttt{C0}'s response, we see that has came to the conclusion that \texttt{C1} is trolling and his \emph{response strategy} is frustrate the trolling event by ignoring the malicious troll's intentions.

\noindent\textbf{Example 2.}
\vspace{-0.1mm}
\begin{itemize}[noitemsep,nolistsep]
  \small{\item[\texttt{C0:}] What do you mean look up ?:( I don't see anything lol}
  \begin{itemize}[noitemsep,nolistsep]
    \small{\item [\texttt{\textbf{C1}}:] \emph{Look up! Space is cool! :)}}
    \begin{itemize}[noitemsep,nolistsep]
      \small{\item[\texttt{C0:}] why must you troll me :(}
      \small{\item[\texttt{C2}:] Keep going, no matter how many times you say it, he will keep asking}
    \end{itemize}
  \end{itemize}
\end{itemize}

In this example, we hypothesize that \texttt{C0} is requesting some information and \texttt{C1} is given an answer that is unfit to \texttt{C0}'s' request. We do so based on the last \texttt{C0}'s comment; \texttt{CO} is showing disappointment or grievance. Also, we deduct that \texttt{C1} is trying to deceive \texttt{C0}, therefore, \texttt{C1}'s comment is a trolling event. This is a trolling event whose \emph{intention} is to purposely convey false information, and that \emph{hiding} its intentions. As for the response, in the last \texttt{C0}'s comment, he has finally realized or \emph{interpreted} that \texttt{C1}'s real intentions are deceiving and since his comment shows a ``sad emoticon'' his reply is emotionally, with aggravation, so we say that \texttt{CO} got \emph{engaged}. \texttt{C2} on the other hand, acknowledges the malicious and play along with the troll.

Given these examples, address the task of predicting the four aspects of a trolling event based on the methodology described in the next section.


\section{Corpus and Annotations}

We collected all available comments in the stories from \textbf{Reddit}\footnote{https://www.reddit.com/} from August 2015. \footnote{Reddit user \emph{Stuck\_In\_the\_Matrix} downloaded all comments from Reddit's api from which we selected a subset.} Reddit is popular website that allows registered users (without identity verification) to participate in forums specific a post or topic. These forums are of they hierarchical type, those that allow nested conversation, where the children of a comment are its direct response. To increase recall and make the annotation process feasible we created an inverted index with Lucene \footnote{https://lucene.apache.org/} and queried for comments containing the word \texttt{troll} with an edit distance of 1, to include close variations of this word. We do so inspired by the method by \cite{xu2012learning} to created a bullying dataset, and because we hypothesize that such comments will be related or involved in a trolling event. As we observed in the dataset, people use the word \emph{troll} in many different ways, sometimes it is to point out that some used is indeed trolling him or her or is accusing someone else of being a troll. Other times, people use the term, to express their frustration or dislike about a particular user, but there is no trolling event. Other times, people simple discuss about trolling and trolls, without actually participating or observing one directly. Nonetheless, we found that this search produced a dataset in which 44.3 \% of the comments directly involved a trolling event. Moreover, as we exposed our trolling definition, it is possible for commentators in a conversation to believe that they are witnessing a trolling event and respond accordingly even where there is none. Therefore, even in the comments that do not involve trolling, we are interested in learning what triggers users interpretation of trolling where it is not present and what kind of response strategies are used. We define as a suspected trolling event in our dataset a comment in which at least one of its children contains the word \emph{troll}.

With the gathered comments, we reconstructed the original conversation trees, from the original post, the root, to the leaves, when they were available\footnote{We removed the comments whose text had been deleted, or the user is marked as \texttt{[deleted]}} and selected a subset to annotated. For annotation purposes, we created snippets of conversations as the ones shown in \textbf{Example 1} and \textbf{Example 2} consisting of the parent of the suspected trolling event, the suspected trolling event comment, and all of the direct responses to the suspected trolling event. We added an extra constraint that the parent of the suspected trolling event should also be part of the direct responses, we hypothesize that if the suspected trolling event is indeed trolling, its parent should be the object of its trolling and would have a say about it. We recognize that this limited amount of information is not always sufficient to recover the original message conveyed by all of the participants in the snippet, and additional context would be beneficial. However, the trade off is that snippets like this allow us to make use of Amazon Mechanical Turk (AMT) to have the dataset annotated, because it is not a big burden for a ``turker'' to work on an individual snippet in exchange for a small pay, and expedites the annotation process by distributing it over dozens of people. Specifically, for each snippet, we requested three annotators to label the four aspects previously described. Before annotating, we set up a qualification test along with borderline examples to guide them in process and align them with our criteria. The qualification test turned out to be very selective since only 5\% of all of the turkers that attempted it passed the exam. Our dataset consists of 1000 conversations with 5868 sentences and 71033 tokens. The distribution over the classes per trolling aspect is shown in the table \ref{tab:results} in the column ``Size''.

\textbf{Inter-Annotator Agreement}. Due to the subjective nature of the task we did not expected perfect agreement. However, we obtained substantial inter-annotator agreement as we measured the fleiss-kappa statistic \cite{fleiss1973equivalence} for each of the trolling aspects: Intention: 0.578,  Intention Disclosure: 0.556, Interpretation: 0.731 and Response 0.632. After inspecting the dataset, we manually reconciled aspects of the threads that found no majority on the turkers annotation and verified and corrected consistency on the four tasks on each thread.

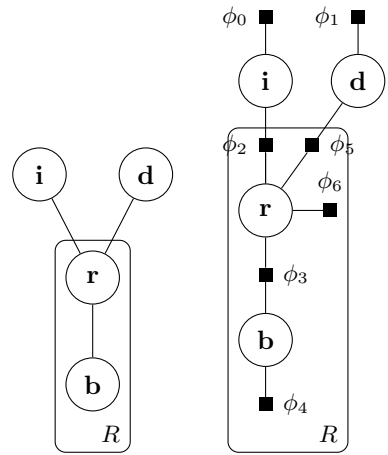
\begin{figure}[ht]
  \begin{center}
    \begin{tabular}{cc}
      \begin{tikzpicture}

  \node[latent]                            (r) {$\mathbf{r}$};
  \node[latent, above=of r, xshift=-0.7cm, yshift=-0.35cm] (i) {$\mathbf{i}$};
  \node[latent, above=of r, xshift=0.7cm, yshift=-0.35cm]  (d) {$\mathbf{d}$};
  \node[latent, below=of r, yshift=0.3cm] (b) {$\mathbf{b}$};

  \edge [-] {i,d} {r} ; %
  \edge [-] {r} {b} ; %

  \plate {rb} {(r)(b)} {$R$} ;

\end{tikzpicture} &
      \begin{tikzpicture}

  
  \node[latent] (i) {$\mathbf{i}$};
  \factor[above=of i] {p0} {left:$\phi_0$} {} {};
  \node[latent, right=of i, xshift=-0.5cm] (d) {$\mathbf{d}$};
  \factor[above=of d] {p1} {left:$\phi_1$} {} {};
  \factor[below=of i] {p2} {left:$\phi_2$} {} {};
  \factor[below=of d,xshift=-0.6cm] {p5} {right:$\phi_5$} {} {};
  \node[latent, below=of i] (r) {$\mathbf{r}$};
  \node[factor] [label=$\phi_6$, right=of r] (p6){};
  \factor[below=of r] {p3} {right:$\phi_3$} {} {};
  \node[latent, below=of r] (b) {$\mathbf{b}$};
  \factor[below=of b] {p4} {right:$\phi_4$} {} {};

  \edge [-] {p0} {i} ; %
  \edge [-] {p1} {d} ;
  \edge [-] {i} {p2} ;
  \edge [-] {p2} {r} ;
  \edge [-] {p2} {r} ;
  \edge [-] {r} {p3} ;
  \edge [-] {p3} {b} ;
  \edge [-] {b} {p4} ;
  \edge [-] {d} {p5} ;
  \edge [-] {r} {p6} ;
  \edge [-] {p5} {r} ;


  \plate {rb} {(r)(b)(p6)(p3)(p4)(p5)(p2)} {$R$} ;

\end{tikzpicture}
    \end{tabular}
  \end{center}
  \caption{Trolling tasks modeled as a (conditional) probabilistic graphical model (left). Factor graph showing all cliques or direct interactions in the model (right). $R$ is the number of direct responses to the suspected trolling comment.} \label{figure:pgms}
\end{figure}


\section{Trolling Events Prediction}

In this section we propose to solve the following problem: given a comment in a conversation, suspected to a trolling event, it's parent comment and all it's direct responses, we aim to predict the suspected comment \texttt{I}: \emph{intention}, its \texttt{D}: \emph{intention disclosure} and from the responses point of view, for each response comment the \texttt{R}: \emph{interpretation} of the suspected troll comment's intentions, and identify its \texttt{B}: \emph{response strategy}. This problem can be seen as a multi-task prediction. To do so, we split the dataset into training and testing sets using a 5-fold cross validation setup.

\subsection{Feature Set}
For prediction we define two sets of features, a basic and an enhanced dataset, extracted from each of the comments in the dataset. The features are described below.

\subsubsection{Basic Feature Set}
\noindent\textbf{N-gram features}. We encode each unigram and bigram collected from the training comments a binary feature. In a similar manner, we include the unigram and bigram along with their POS tag as in \cite{xu2012learning}. To extract these features we used the most current version of the Stanford CoreNLP \cite{manning-EtAl:2014:P14-5}. Each token's \noindent\textbf{Lemmas} as in \cite{xu2012fast} as a binary feature.

\noindent\textbf{Harmful Vocabulary}. In their research on bullying \cite{nitta2013detecting} identified a small set of words that are highly offensive. We encode them as well as binary features.

\noindent\textbf{Emotions Synsets}. As in \cite{xu2012fast} we extracted all lemmas associated with each of \emph{Synsets} extracted from WordNet \cite{miller1995wordnet} from these emotions: anger, embarrassment, empathy, fear, pride, relief and sadness. As well all the synonyms from these emotions extracted from the dictionary. Also, 

\subsubsection{Enhanced Feature Set}

\noindent\textbf{Emoticons}. Reddit's comments make extensive use of emoticons, we argue that some emoticons are specially used in trolling events and to express a variety of emotions, which we hypothesize would be useful to identify a comments intention, interpretation and response. For that we use the emoticon dictionary \cite{hogenboom2015exploiting} and we set a binary feature for each emoticon that is found in the dictionary.

\noindent\textbf{Sentiment Polarity}. Using a similar idea, we hypothesize that the overall comment emotion would be useful to identify the response and intention in a trolling event. So, we apply the Vader Sentiment Polarity Analyzer \cite{hutto2014vader} and include a four features, one per each measurement given by the analyzer: positive, neutral, negative and a composite metric, each as a real number value.

\noindent\textbf{Subjectivity Lexicon}. From the MPQA Subjective Lexicon \cite{wilson2005recognizing} we include all tokens that are found in the lexicon as binary features. This lexicon was created from a news domains, so the words in it don't necessarily align with the informal vocabulary used in Reddit, but, there are serious Reddit users that use proper language and formal constructions. We believe that these features will allow us to discriminate formal comments from being potentially labeled as trolling events, which tend to be vulgar.

\noindent\textbf{Swearing Vocabulary}. We manually collected 1061 swear words and short phrases from the internet, blogs, forums and smaller repositories. The informal nature of this dictionary resembles the type of language used by flaming trolls and agitated responses, so we encode a binary feature for each word or short phrase in a comment if it appears in the swearing dictionary.

\noindent\textbf{Framenet}. Following \cite{hasan2014you} use of FrameNet, we apply the Semaphore Parser \cite{das2014frame} to each sentence in every comment in the training set, and construct three different binary features: every frame name that is present in the sentence, the frame name a the target word associated with it, and the argument name along with the token or lexical unit in the sentence associated with it. We argue that some frames are especially interesting from the trolling perspective. For example, the frame ``Deception\_success'' precisely models one of the trolling models, and we argue that these features will be particularly to identify trolling events in which semantic and not just syntactic information is necessary.

\noindent\textbf{Politeness Queues}. \cite{danescu2013computational} identified queues that signal polite and impolite interactions among groups of people collaborating online. Based on our observations of trolling examples, it is clear that flaming troll and engaged or emotional responses would use impolite queues. On the contrary, neutralizing and frustrating responses to troll avoid falling in confrontation and their vocabulary tends to be more polite. So use these queues as binary features as they appear in the comments in consideration.

\subsection{Baseline System}

The most na\"ive approach is to consider each of the four tasks as an independent classification problem. Such system would be deprived from the other's tasks information that we've mentioned is strictly necessary to make a correct prediction of the response strategy. Instead, as our baseline we follow a pipeline approach, using the tasks oder: \texttt{I}, \texttt{D}, \texttt{R} and \texttt{B}, so that each of the subsequent subtasks' feature set is extended with a feature per each of previously computed subtasks. We argue that this setup is a competitive baseline, as can be verified in the results table \ref{tab:results}. For the classifier in the pipeline approach we choose a log-linear model, a logistic regression \footnote{We use \emph{scikit-learn}\cite{scikit-learn} implementation for all baseline experiments.} classifier. In addition to logistic regression, we tried the generative complement of logistic regression, na\"ive bayes and max-margin classifier, a support vector machine, but their performance was not superior to the logistic regression. It is noteworthy to mention that the feature set used for the \emph{intention} predict is the combined features sets of the suspected troll comment as well as its parent. We do so in all of our experiments the learner can take advantage of the conversation context.

\subsection{Joint Models}

The nature of this problem makes the use of a joint model a logical choice. Among different options for joint inference, we choose a (conditional) probabilistic graphical model (henceforth PGM)\cite{koller2009probabilistic} because, in contrast to ILP formulations, has the ability to learn parameters and not just impose hard constraints. Also, compared to Markov Logic Networks \cite{richardson2006markov}, a relatively recent formulation that combines logic and Markov Random Fields, PGMs in practice have proved to be more scalable, even though, inference in general models is shown to be intractable. Finally, we are also interested in choosing a PGM because it allow to directly compare the strength of joint inference with the baseline, because the our model is a collection of logistic regressors trained simultaneously.

A conditional random field factorizes the conditional probability distribution over all possible values of the query variables in the model, given a set of observations as in equation \ref{eq:1}. In our model, the query variables are the four tasks we desire to predict, $\mathbf{y}$ and the observations is their combined feature sets $\mathbf{x}$. Each of the factors $\Psi_a$ in this distribution is a log-linear model as in equation \ref{eq:2} and represents the probability distribution of the clique of variables $\mathbf{y}$ in it, given the observation set $\mathbf{x}$. This is identical to the independent logistic regression model described in the baseline, except for the fact that all variables or tasks are consider a the same time. To do so, we add additional factors that connect task variables among them, permitting the flow of information of one task to the other.

Specifically, our model represent each task with a random variable, shown in figure \ref{figure:pgms} (left), represented by the circles. The plate notation that surrounds variables $r$ and $b$ indicates that there will as many variables $r$ and $b$ and edges connecting them to $i$ as the number of responses in the problem snippet. The edges connecting $i$ and $d$ with $r$ attempts to model influence of these two variables on the response, and how this information is passed along to the response strategy variable $b$. Figure \ref{figure:pgms}(right) explicitly represents the cliques in the underlying factor graph. We can see that there are unary factors, $\phi_0$, $\phi_1$, $\phi_4$ and $\phi_6$, that model the influence of the observation features over their associated variables, just as the logistic regression model does. Factors $\phi_2$ models the interaction between variables $i$ and $r$, $\phi_5$ the interaction between variables $d$ and $r$ and $\phi_3$ models the interactions between variables $r$ and $b$, using a log-linear model over the possible values of the pair of variables in that particular clique.

Due to the size of the model, we are able to perform exact inference at train and test time. For parameter learning we employ limited memory lbfgs optimizer \cite{byrd1995limited} as we provide the cost function and gradient based on the equations described in \cite{sutton2006introduction}.

\textbf{2 pass Model} A hybrid mode that we experiment with is model that performs joint inference on three tasks: \texttt{I:} intention, \texttt{D:} intention disclosure and \texttt{R:} responders' intention interpretation. The remaining task \texttt{B:} response strategy is performed in a second step, with the input the other three tasks. We do so because we observed in our experiments that the close coupling between the first three tasks allow them to perform better independently of the response strategy, as we will elaborate in the results section.

\begin{equation}\label{eq:1}
p(\textbf{y}|\textbf{x};\theta)=\frac{1}{Z(\textbf{x})}\prod_{a=1}^{A}\Psi_a(\textbf{y}_a,\textbf{x}_a)
\end{equation}

\begin{equation}\label{eq:2}
\Psi_a(\textbf{y}_a,\textbf{x}_a)=exp\left\{\sum_{k=1}^{K(A)}\theta_{ak}f_{ak}(\textbf{y}_a,\textbf{x}_a)\right\}
\end{equation}

\section{Evaluation and Results}

We perform 5-fold cross validation on the dataset. We use the first fold to tune the parameters and the remaining four folds to report results. The system performance is measured using precision, recall and F-1 as shown in table \ref{tab:results}. The left side of the table, reports results obtained using the basic feature set, while the right side does so on the enhanced feature set. In order to maintain consistency folds are created based on the threads or snippets and for the case of the baseline system, all instances in the particular fold for task in consideration are considered independent of each other. On the table, rows show the classes performance for each of the tasks, indicated by a heard with the task name. For the response strategy we present results for those class values that are at least 5\% of the total distribution, we do so, because the number of labeled instances for this classes is statistically insignificant compared to the majority classes. 

\begin{table*}[!th]
\begin{small}
\centering 
{\resizebox{\textwidth}{!}{
\begin{tabular} {|l|p{3.5mm}p{3.5mm}p{4.5mm}| p{3.5mm}p{3.5mm}p{4.5mm}| p{3.5mm}p{3.5mm}p{4.5mm}| p{3.5mm}p{3.5mm}p{4.5mm}| p{3.5mm}p{3.5mm}p{4.5mm}| p{3.5mm}p{3.5mm}p{4.5mm}|p{3.5mm}|} 
\hline
& \multicolumn{9}{c|}{Basic Feature Set} & \multicolumn{9}{c|}{Enhanced Feature Set} &  \multicolumn{1}{c|}{Size}\\ 
\cline{2-20}
& \multicolumn{3}{c|}{Baseline} & \multicolumn{3}{c|}{Full Joint} & \multicolumn{3}{c|}{Joint + 2nd Pass} & \multicolumn{3}{c|}{Baseline} & \multicolumn{3}{c|}{Full Joint} & \multicolumn{3}{c|}{Joint + 2nd Pass} & \multicolumn{1}{c|}{-} \\ 
\cline{2-20}
Experiment/Class  & P & R & F1  & P & R & F1  & P & R & F1  & P & R & F1  & P & R & F1  & P & R & F1 & \% \\ 
\hline
\multicolumn{20}{l}{\textbf{I: Intention}} \\
\hline
No trolling & 64.0& 72.8& 68.2& 91.4& 29.6& 44.6& 91.6& 89.0& \textbf{90.2}& 64.2& 72.2& 67.8& 92.4& 33.8& 49.2& 89.2& 90.8& 90.0& 55.7\\
Trolling & 57.2& 50.4& 53.4& 50.0& 98.2& 66.4& 87.8& 93.0& \textbf{90.4}& 56.4& 50.4& 52.8& 51.6& 98.2& 67.4& 88.8& 89.4& 89.2& 41.7\\
Playing & 0.0& 0.0& 0.0& 0.0& 0.0& 0.0& 80.0& 54.0& \textbf{63.6}& 0.0& 0.0& 0.0& 0.0& 0.0& 0.0& 80.0& 50.0& 60.8& 2.6\\
\hline
\multicolumn{20}{l}{\textbf{D: Disclosure}} \\
\hline
None& 67.0& 78.0& 71.8& 80.0& 91.4& 85.2& 89.6& 90.6& 90.0& 67.6& 78.0& 72.2& 81.6& 80.2& 80.8& 89.4& 91.0& \textbf{90.2}& 55.4\\
Hidden & 0.0& 0.0& 0.0& 80.0& 22.0& 34.0& 80.8& 62.0& 69.2& 0.0& 0.0& 0.0& 80.0& 8.6& 15.2& 75.8& 67.2& \textbf{69.8}& 8.4\\
Exposed & 56.6& 55.4& 55.4& 80.0& 77.4& 78.8& 84.2& 88.4& 86.0& 57.2& 56.8& 56.4& 68.2& 84.0& 75.2& 86.0& 86.2& \textbf{86.2}& 36.2\\
\hline
\multicolumn{20}{l}{\textbf{R: Interpretation}} \\
\hline
No trolling & 71.0& 63.2& 66.6& 84.6& 36.2& 50.4& 88.2& 76.6& 82.0& 71.0& 63.6& 67.0& 87.0& 41.2& 55.8& 85.4& 81.6& \textbf{83.0}& 38.9\\
Trolling & 77.2& 84.0& 80.6& 69.6& 96.8& 81.0& 85.8& 94.6& \textbf{90.0}& 77.4& 83.8& 80.4& 71.6& 96.8& 82.4& 87.8& 92.2& \textbf{90.0}& 60.0\\
Playing & 0.0& 0.0& 0.0& 0.0& 0.0& 0.0& 0.0& 0.0& 0.0& 0.0& 0.0& 0.0& 0.0& 0.0& 0.0& 0.0& 0.0& 0.0& 1.1\\
\hline
\multicolumn{20}{l}{\textbf{B: Response}} \\
\hline
Engage & 30.6& 31.6& 30.4& 0.0& 0.0& 0.0& 34.2& \textbf{31.8}& 31.8& 29.4& 32.0& 29.4& 0.0& 0.0& 0.0& 33.6& 30.4& 30.0& 12.4\\
False Accusation & 20.4& 24.8& 22.0& 10.0& 0.6& 1.2& 24.2& 30.8& \textbf{26.8}& 21.8& 27.0& 23.8& 20.0& 0.6& 1.2& 23.4& 28.0& 25.2& 14.4\\
Neutralize & 26.0& 39.2& 31.0& 18.8& 98.6& 31.6& 28.6& 40.6& \textbf{33.4}& 26.4& 37.6& 30.8& 19.4& 98.0& 32.0& 26.2& 38.0& 30.6& 15.7\\
Normal & 41.4& 58.6& 48.4& 46.6& 41.0& 43.6& 41.8& 60.8& \textbf{49.4}& 40.2& 58.6& 48.0& 45.8& 44.6& 45.2& 40.2& 60.0& 47.8& 18.4\\
Frustrate & 15.0& 14.8& \textbf{14.4}& 0.0& 0.0& 0.0& 13.0& 12.0& 11.8& 15.2& 14.2& 14.2& 20.0& 1.0& 2.0& 13.2& 12.0& 12.2& 9.9\\
Imaginary Bite & 22.2& 8.8& 11.6& 0.0& 0.0& 0.0& 30.6& 9.2& \textbf{13.0}& 20.4& 8.8& 11.4& 0.0& 0.0& 0.0& 22.4& 9.2& 12.4& 5.8\\
Bite Attempt & 17.6& 13.2& 15.0& 0.0& 0.0& 0.0& 19.4& 14.0& \textbf{15.6}& 13& 10.2& 11.4& 0.0& 0.0& 0.0& 16.0& 11.8& 13.4& 9.6\\

\hline
\end{tabular}}}
\end{small}
\captionsetup{justification=centering}
\caption{Prediction Results for the four aspects of trolling: Intention, Intentions Disclosure, Interpretation, and Response strategy. Three models are evaluated: a logistic regression classifier: Baseline, a four-tasks CRF: Full Joint, and a two steps process: three-tasks CRF followed by the Response Strategy prediction tasks given the the outcome of the CRF} \label{tab:results}
\end{table*}

\subsection{Results Discussion}

From the result table \ref{tab:results}, we observe that hybrid model significantly outperform the baseline, by more than 20 points in intention and intention disclosure prediction. For the response strategy, it is clear that none of the systems offer satisfying results; this showcases the difficult of such a large number of classes. Nonetheless, the hybrid model outperforms the fully joint model and baseline in all but one the response strategy classes. However, the differences are far less impressive as in the other tasks. It is surprisingly; that the full joint model did not offered the best performance. One of the reasons behind this is that intention, intentions disclosure and interpretation tasks are hurt by the difficulty of learning parameters that maximize the response strategy, this last task drags the other three down in performance. Another reason is that, the features response strategy is not informative enough to learn the correct concept, and due to the joint inference process, all tasks receive a hit. Also, it is counter-intuitive that the augmented set of features did not outperform in all tasks but in intentions disclosure and interpretation, and just by a small margin. A reason explaining this unexpected behavior is that the majority of enhanced features are already represented in the basic feature set by means of the unigrams and bigrams, and the Framenet and Sentiment features are uninformative or redundant. Lastly, we observe that for interpretation category, none of systems were able to predict the ``playing'' class. This is because of the relative size of the number of instances annotated with that value, 1\% of the entire dataset. We hypothesize those instances labeled by the annotators, of which a majority agreed on, incorrectly selected the playing category instead of the trolling class, and that, at the interpretation level, one can only expect to reliably differentiate between trolling and trolling.

\section{Related Work}

In this section, we discuss related work in the areas of trolling, bullying and politeness, as they intersect in their scope and at least partially address the problem presented in this work.

\cite{mihaylov2015finding} address the problem of identifying manipulation trolls in news community forums. The major difference with this work is that all their predictions are based on meta-information such as number of votes, dates, number of comments and so on. There is no NLP approach to the problem and their task is limited to identifying trolls. \cite{bishop2013effect} and \cite{bishop2014representations} elaborate a deep description of the trolls personality, motivations, effects on the community that trolls interfere and the criminal and psychological aspects of trolls. Their main focus are flaming trolls, but have no NLP insights do not propose and automated prediction tasks as in this work. In a networks related framework \cite{kumar2014accurately} and \cite{guha2004propagation} present a methodology to identify malicious individuals in a network based solely on the network's properties. Even though they offer present and evaluate a methodology, their focus is different from NLP. \cite{cambria2010not} proposes a method that involves NLP components, but fails to provide a evaluation of their system. Finally, \cite{xu2012learning} and \cite{xu2012fast} address bullying traces. That is self reported events of individuals describing being part of bullying events, but their focus is different from trolling event and the interactions with other participants.

\section{Conclusion and Future Work}

In this paper we address the under-attended problem of trolling in Internet forums. We presented a comprehensive categorization of trolling events and defined a prediction tasks that does not only considers trolling from the troll's perspective but includes the responders to the trolls comment. Also, we evaluated three different models and analyzed their successes and shortcomings. Finally we provide an annotated dataset which we hope will be useful for the research community. We look forward to investigate trolling phenomena in larger conversations, formalize the concepts of changing roles among the participants in trolling events, and improve response strategy performance.

\appendix

\section{Response Strategy Definitions}
\begin{enumerate}[noitemsep,nolistsep]
  \item Normal: unemotional and standard response to a comment that is not a trolling event.
  \item Bite Attempt: response comment with malicious intentions to a parent comment that is not a trolling event
  \item Imaginary Bite: reply with aggravation thinking that he/she is being trolled but without the intention of trolling back.
  \item False accusation: comment criticizing or denouncing malicious intentions of parent's comment, but the parent comment has no malicious intentions.
  \item Frustrate: comment that acknowledges the parent's malicious or playful intentions but refuses to engage or fall in the provocation as it gives no importance to them.
  \item Neutralize: comment that acknowledges the malicious or playful intentions and tries to minimize or criticize them
  \item Counter Trolling: comment that acknowledges the malicious or playful intentions and attempts to troll back with a trolling event
  \item Praise: comment that acknowledges the malicious or playful intentions, but positively recognize the troll's ingenuity or ability
  \item Engage: comment that acknowledges the malicious or playful intentions and falls in the provocation, giving a discomposed response. 
  \item Aggravation: comment that acknowledges the non-malicious or playful intentions, but they are received emotionally
  \item Confrontation: annoyed response based on the belief of malicious intentions when the original intentions are playful.
  \item Failed: comment response in which the responder doesn't realized the malicious or playful intentions, but does not fall in the provocation. We say that the troll failed.
  \item Bite: comment response that falls in the malicious intentions of a deceiving trolling event or a upset reply to a flaming trolling event
  \item Follow: comment response in which the responder understand the the malicious or playful intentions and plays along with the troll.

\end{enumerate}

\bibliographystyle{acl2016}
\bibliography{acl2016}

\end{document}